\documentclass[conference]{IEEEtran}
\IEEEoverridecommandlockouts
\usepackage{cite}
\usepackage{amsmath,amssymb,amsfonts}
\usepackage{algorithmic}
\usepackage{graphicx}
\usepackage{textcomp}
\usepackage{xcolor}
\usepackage{subcaption}
\usepackage{tikz}
\usepackage{booktabs}
\usepackage{multirow}

\usepackage{hyperref} 
\hypersetup{backref=true,       
    pagebackref=true,               
    hyperindex=true,                
    colorlinks=true,                
    breaklinks=true,                
    urlcolor= black,                
    linkcolor= blue,                
    bookmarks=true,                 
    bookmarksopen=false,
    filecolor=black,
    citecolor=blue,
    linkbordercolor=blue
}

\newcommand\copyrighttext{%
  \footnotesize \textcopyright 2020 IEEE. Personal use of this material is permitted.  Permission from IEEE must be obtained for all other uses, in any current or future media, including reprinting/republishing this material for advertising or promotional
  purposes, creating new collective works, for resale or redistribution to servers or lists, or reuse of any copyrighted component of this work in other works.}
\newcommand\arxivcopyrightnotice{%
\begin{tikzpicture}[remember picture,overlay]
\node[anchor=south,yshift=10pt] at (current page.south) {\fbox{\parbox{\dimexpr\textwidth-\fboxsep-\fboxrule\relax}{\copyrighttext}}};
\end{tikzpicture}%
}

\def\BibTeX{{\rm B\kern-.05em{\sc i\kern-.025em b}\kern-.08em
    T\kern-.1667em\lower.7ex\hbox{E}\kern-.125emX}}

\usepackage{array}
\newcolumntype{P}[1]{>{\centering\arraybackslash}m{#1}}

\begin{document}
\bibliographystyle{IEEEtran}

\title{Quantisation and Pruning for Neural Network \\ Compression and Regularisation
\thanks{This work is based on the research supported in part by the National Research Foundation of South Africa (Grant Numbers: 118075 and 117808).}
}

\author{\IEEEauthorblockN{Kimessha Paupamah}
\IEEEauthorblockA{\textit{School of Computer Science} \\
\textit{and Applied Mathematics} \\
\textit{University of the Witwatersrand}\\
Johannesburg, South Africa \\
kimessha.paupamah1@students.wits.ac.za}
\and
\IEEEauthorblockN{Steven James}
\IEEEauthorblockA{\textit{School of Computer Science} \\
\textit{and Applied Mathematics} \\
\textit{University of the Witwatersrand}\\
Johannesburg, South Africa \\
steven.james@wits.ac.za}
\and
\IEEEauthorblockN{Richard Klein}
\IEEEauthorblockA{\textit{School of Computer Science} \\
\textit{and Applied Mathematics} \\
\textit{University of the Witwatersrand}\\
Johannesburg, South Africa \\
richard.klein@wits.ac.za}
}

\maketitle

\arxivcopyrightnotice

\IEEEpubidadjcol

\begin{abstract}
Deep neural networks are typically too computationally expensive to run in real-time on consumer-grade hardware and low-powered devices. In this paper, we investigate reducing the computational and memory requirements of neural networks through network pruning and quantisation. We examine their efficacy on large networks like AlexNet compared to recent compact architectures: ShuffleNet and MobileNet. Our results show that pruning and quantisation compresses these networks to less than half their original size and improves their efficiency, particularly on MobileNet with a 7$\times$ speedup. We also demonstrate that pruning, in addition to reducing the number of parameters in a network, can aid in the correction of overfitting.
\end{abstract}

\section{Introduction}

Designing deep and complex neural networks is common practice for effective performance on various applications, particularly visual tasks like image classification \cite{liu2017surveydeep}. As neural networks become larger and deeper, more computational resources are required to train and store them, making it increasingly more difficult to deploy these networks on the consumer-grade hardware, mobile and embedded devices in use today. In addition to requiring a large amount of computational resources, deep neural networks take up tremendous amounts of energy, leaving a large carbon footprint. For example, \cite{strubell1906energy} show that training certain deep natural language processing (NLP) models can result in as much CO$_2$ emissions as five cars in their lifetime. Small, powerful neural networks would help overcome these problems.

We can employ methods of neural network compression to obtain small and efficient neural networks which consume much less energy and can consequently be deployed to devices with limited computing capabilities. Large neural networks often contain many redundant parameters that have no impact on the network \cite{denil2013predicting}. Removing or \emph{pruning} these redundant parameters result in networks with lower complexity. Quantisation is a further approach to reduce the size of neural networks by lowering the number of bits required to represent parameters. Another approach to obtain small networks is to directly build smaller, efficient network architectures. These compact architectures, like MobileNet \cite{howard2017mobilenets} and ShuffleNet \cite{zhang2018shufflenet}, perform computationally efficient operations and produce networks that are small in size, which makes them easy to deploy on mobile and embedded devices.

These approaches of obtaining smaller networks fall into two categories: either reduce the size of large networks, or directly train small, compact networks. The aim of this work is to compare these two approaches and examine their sensitivity to compression techniques in terms of accuracy, size, and inference time. Furthermore, we examine the effects of pruning as a means of correcting overfitting. We conduct our experiments on the CIFAR-10 \cite{krizhevsky2014cifar} and FashionMNIST \cite{xiao2017fashion} datasets. We find that overfitted networks benefit from pruning, and that compact architectures outperform large, compressed networks.


\section{Background}
This section aims to provide the background necessary for understanding neural network compression and the methods thereof. We give a brief overview of neural networks and convolutional neural networks, followed by a discussion on separable convolutions then network pruning and quantisation.

\subsection{Neural Networks}

A typical feedforward neural network is composed of a number of artificial neurons which are organised into layers, the first being the input layer while the last being the output layer. The layers between are the hidden layers, which form the capacity of the network. Neurons that reside in a layer are linked to neurons in the subsequent layer by weighted connections. These weights form the \emph{parameters} of the neural network.

An artificial neuron performs some mathematical operation, usually by taking the dot product of the input connections and passing it through some activation function to \emph{fire} an output through an output connection, consequently weighting the connection \cite{Goodfellow-et-al-2016}. These output connections form the input connections for neurons in the subsequent layer and so the output is propagated to other connected neurons to give the final output of the network. This process is called a forward pass or forward propagation.

A neural network is trained by learning its parameters. A common method for learning the parameters of a neural network is through backpropagation \cite{rumelhart1988learning}. First, a forward pass of the network occurs to predict the final output, and a loss function is used to measure the error of the prediction. Optimisation techniques like gradient descent are used to minimise the loss and find optimal weights for the connections; however, the partial derivatives of the loss function are required. Backpropagation is a method used to compute these partial derivatives by propagating the loss from the output layer back through the network to the input layer and computing how each neuron contributes to the loss.

\subsection{Convolutional Neural Networks}
Convolutional neural networks are similar to feedforward neural networks, except their layers are composed of convolutional layers which have a height, width and depth. A convolutional layer performs convolutions with input to extract features, and so has of a set of filters (or weights) which are learnt during training \cite{Goodfellow-et-al-2016}. Mathematically, a convolution operation with an image can be described by
\begin{equation}
    (\mathbf{I}*\mathbf{K})[i,j]=\sum_{p=0}^{m-1}\sum_{q=0}^{n-1}\mathbf{I}[i-p,j-q]\mathbf{K}[p,q]
\end{equation}
for an image $\mathbf{I}$, of size $M \times N$, and kernel $\mathbf{K}$, of size $m \times n$. The image is convolved with a kernel by sliding the kernel across each pixel in the image and taking the dot product of the kernel elements and the pixel values aligned with the kernel. We convolve a stack of kernels, or filter, of the same depth as the number of colour channels in the image. A number of filters can be convolved with the image, each producing an output channel.

The convolutional layer arranges neurons in a three-dimensional grid. Neurons in the convolutional layer are only connected to a local region of the input. This region is called the receptive field of a neuron and is the same size as the filter used. These neurons work similarly to neurons in feedforward networks by convolving the filter with a local region of the input, then passing it through an activation function to give output channels which are stacked together to form a feature map. Other popular layers in a convolutional neural networks include Batch Normalisation \cite{ioffe2015batch} and  Dropout \cite{srivastava2014dropout} layers.

\subsection{Separable Convolutions}
Consider an input size of $w_{in} \times h_{in} \times d$, convolved with a filter of size $N \times k \times k \times d$, to give a feature map of size $w_{out} \times h_{out} \times N$. This standard convolution has a computational cost of $w_{in} \times h_{in} \times d \times N \times k \times k$. We look at reducing this computational cost with depthwise separable convolutions and group convolutions.

\subsubsection{Depthwise Separable Convolutions}
\begin{figure}[b!]
\begin{center}
\includegraphics[width=6cm,height=7cm]{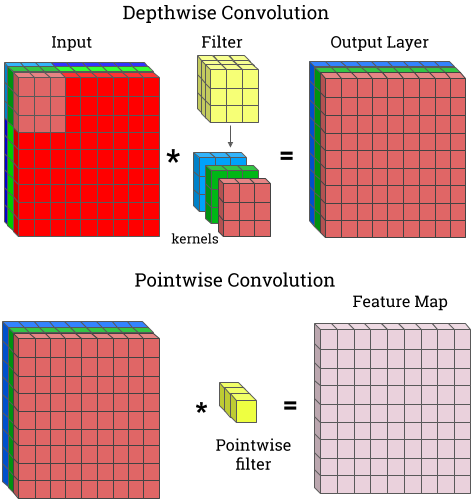}
\end{center}
\caption{Example of a depthwise separable convolution on an RGB image with $N=1$}
\label{fig:depthsepconv}
\end{figure}

Depthwise separable convolutions \cite{sifre2014rigid} first perform a depthwise convolution followed by a pointwise convolution. Depthwise convolutions perform convolutions along the input channels separately. Each filter is sliced into $d$ separate $k \times k$ kernels, and each kernel is then convolved with its own input channel. The output of each convolution is stacked together to form an output layer of size $w_{out} \times h_{out} \times d \times N$. To transform this output layer to a single feature map of size $w_{out} \times h_{out} \times N$, a pointwise convolution is performed by convolving this layer with a $1\times1 \times d$ filter $N$ times. This process is illustrated in Fig. \ref{fig:depthsepconv} using one filter ($N=1$). The total computation cost can be calculated as $(w_{in} \times h_{in} \times d \times k \times k) + (w_{in} \times h_{in} \times d \times N)$ which results in a significant reduction in computation.

\subsubsection{Group Convolutions}
\begin{figure}
\begin{center}
\includegraphics[width=6cm,height=3cm]{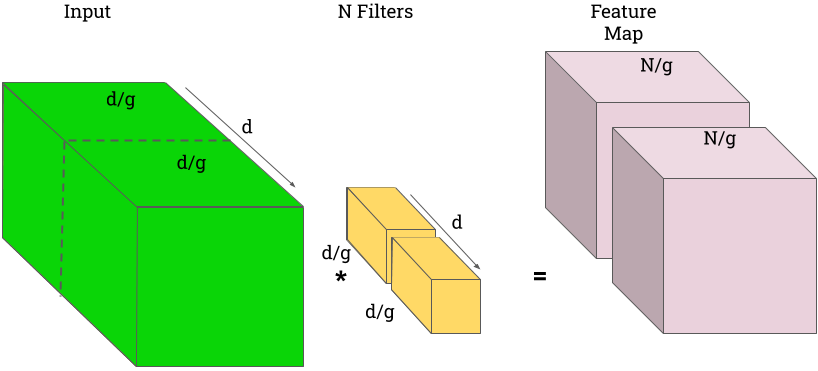}
\end{center}
\caption{Example of a group convolution with $g=2$}
\label{fig:groupconv}
\end{figure}

Group convolutions \cite{krizhevsky2012imagenet} operate by dividing filters into different groups, with each filter group being convolved with a different part of the input layer of a certain depth. A filter can be divided into $g$ separate groups along its depth. This results in $g$ groups, with each group consisting of $N/g$ separate filters of size $k \times k \times d/g$. The same is done to the input layer to get $g$ separate layer groups of $w_{in} \times h_{in} \times d/g$. Each $N/g$ group is convolved with an input layer group of the same depth, to get an output layer of size $w_{out} \times h_{out} \times N/g$. These resulting output layers from each group convolution are stacked together to obtain the resultant feature map. Fig. \ref{fig:groupconv} illustrates this with two groups ($g=2$). The filters and input layer are divided into two groups: the first filter group convolves with the first half of the input, while the second filter group convolves with the second half of the input.
Grouped convolutions reduces the computational cost to $g \times (w_{in} \times h_{in} \times d/g \times N/g)$. Since the convolutions are divided, group convolutions also allow for efficient computation as each convolution can be handled in parallel, on a separate GPU for instance.

\subsection{Network Pruning}
The general procedure for pruning a trained neural network is locating which parameters have no significant impact on the network and then removing those redundant parameters. The network is then retrained after this pruning process so that the remaining parameters in the network are adjusted to compensate for those removed. In this work, we focus on iterative pruning as introduced in \cite{han2015alearning}. This iterative procedure is a three-step method. The first step of the method fully trains the neural network to learn the parameters (or connections). Once the network is fully trained, the second step of the method is to learn which connections in the network are important. These important connections are learnt iteratively, where an iteration involves pruning connections with weights below a threshold and then retraining the network.  The threshold value is found manually, by determining which layers are sensitive to pruning.  After this pruning stage, neurons which have no input or output connections are also removed, and hence all further connections to and from the pruned neuron are removed.  This results in a sparse neural network, with the unimportant connections pruned away and important connections preserved. The final step of this method is to retrain the resulting sparse network. During the retraining stages, the weights are not re-initialised to ensure that gradient descent finds a good solution. This is also computationally cheaper since there is no need to backpropagate through the entire network.

\subsection{Network Quantisation}
Network quantisation is a method of reducing the precision of weights and activations in neural networks by lowering the number of bits to represent these quantities \cite{krishnamoorthi2018quantizing}. This is a quick and efficient way to reduce a network's size without the need for retraining. We can quantise a parameter $x$ according to the mapping
\begin{equation}
    Q(x, \Delta, z) = \text{round}(\frac{x}{\Delta} + z).
\end{equation}
This maps floating-point values to integers, which in turn lowers the number of bits required to represent a parameter. The scale ($\Delta$) indicates the step size of the quantiser, while a floating-point zero maps to an integer $z$, the zero-point, so that zero can be quantised with no error.


\section{Related Work}
Early pruning methods like Optimal Brain Damage \cite{lecun1990optimal} and Optimal Brain Surgeon \cite{hassibi1993second} pruned shallow, fully-connected networks based on saliency values, which indicate the effect of a parameter on the training error. These values are computed by approximating the Hessian matrix which is computationally expensive for the large, deep networks in practice today. Recent works have studied pruning larger and deeper neural networks. An iterative pruning method which results in sparse networks is introduced in \cite{han2015alearning}. ThiNet \cite{luo2017thinet} introduces a filter pruning technique which removes entire filters in convolutional neural networks. The authors propose a method using a least-squares approach to find channels corresponding to unimportant filters. These weak channels in the next layer of a network are found and their corresponding filters in the current layer are pruned away. This thins down the original wider network and reduces the computational cost compared to magnitude pruning. Additionally, no sparsity is introduced into the network, and so the original network structure remains intact. Bayesian Compression \cite{louizos2017learning} introduce a Bayesian approach to pruning. Sparsity inducing priors are used to prune entire weight structures rather than individual parameters. This results in very sparse networks with high compression rates. 

Other means of reducing the number of weights or parameters in neural networks are methods of weight sharing and quantisation. Weight sharing methods allow weights within layers of the network to be shared, while quantisation simply represents weights with a lower number of bits. This leads to a more accelerated network with reduced complexity. HashNets \cite{chen2015compressing} is a neural network architecture that operates by grouping weights together into hash buckets using a hash function. The assignment of weights to connections are determined by a hash function so that all connections grouped to the same hash bucket share the same weight. Deep Compression \cite{han2016deep} employs both weight sharing and quantisation as an additional compression method after pruning networks. This is done by clustering all the weights within a layer, then approximating each weight to the closest cluster centroid, lowering the number of bits needed to store each weight.

There has also been interest in building efficient architectures. Like MobileNet and ShuffleNet, these compact architectures perform several varieties of convolutions to reduce computational cost. GoogLeNet \cite{szegedy2015googlenet} allows for the increase in depth and width of a network while maintaining a constant computational cost. The architecture uses Inception modules which combines convolutions at different scales to help with dimensionality reduction. SqueezeNet \cite{iandola2016squeezenet} designs very small networks with bottleneck layers that squeeze input and expands it afterwards.


\section{Methodology}
We have outlined two approaches of obtaining small networks: reducing the size of large networks through network compression or directly building small and efficient compact architectures. We compare these two opposing ideas to obtain an understanding of each method and which to employ in practice. Our comparison also examines the response of compression on small, compact architectures in an effort to understand how they are impacted. Additionally, we investigate the effects of correcting overfitted networks with pruning, to determine whether pruning can be used as an effective regularisation technique. This section outlines our approach to neural network compression and the overfitting of networks.

\subsection{Datasets and Network Architectures}
We test our experiments on the CIFAR-10 \cite{krizhevsky2014cifar} and FashionMNIST \cite{xiao2017fashion} datasets. We choose to use AlexNet \cite{krizhevsky2012imagenet} as our large network as it contains tens of million parameters and can fit onto the hardware available to us. AlexNet is an eight-layer deep network, containing five convolutional layers followed by three fully-connected layers. For our compact architectures we use the improved state-of-the-art MobileNetV2 \cite{sandler2018mobilenetv2} and ShuffleNetV2 \cite{ma2018shufflenetv2} architectures. The MobileNetV2 architecture falls under the class of MobileNet architectures, and similarly the ShuffleNetV2 architecture falls under the class of ShuffleNet architectures, and so shall be referred to as \emph{MobileNet} and \emph{ShuffleNet} respectively. MobileNet's input layer is fully convolutional, followed by eighteen inverted residual block hidden layers, which perform depthwise separable convolutions. The output layer is a single fully connected layer to perform classification. ShuffleNet's architecture is composed of three stages, each having a repeated stack of inverted residual blocks, which perform pointwise group convolutions with channel shuffling followed by a depthwise convolution, and then another pointwise convolution. The input layer of the network is fully convolutional, while the output layer is a single fully-connected layer. 
These networks were trained from scratch and used as our reference networks in our experiments. They were trained with a batch size of 50 and optimised using stochastic gradient descent with a momentum of 0.9 on both datasets. The learning rate decayed during training, with AlexNet starting with a learning rate of 0.001 while MobileNet and ShuffleNet started with a learning rate of 0.01. Our experiments implemented in PyTorch \cite{paszke2017pytorch} and run them on Nvidia GeForce GTX 1060 Ti and 1080 Ti GPUs. The source code is available online.\footnote{https://github.com/kpaupamah/compression-and-regularisation}

\subsection{Network Compression} 
To compress our networks, we first iteratively prune them. A pruning iteration consists of pruning parameters that have no impact on the network then retraining the resulting sparse network. The pruned parameters are those with the smallest weights, and the number of parameters removed is determined by the network's sensitivity to pruning. 

Our pruned models are further compressed by applying per-channel quantisation \cite{krishnamoorthi2018quantizing}. Per-channel quantisation lowers the bits used to represent parameters along the depth (or channels) of a layer. Applying quantisation results in minor accuracy loss with a smaller network size, and a speedup in terms of training and inference time.

\subsection{Overfitting}
To overfit our networks, we remove all regularisation layers and train our networks until we completely learn our training data. In particular, we remove the Dropout layers from all three networks and remove the BatchNorm layers from both MobileNet and ShuffleNet. Training our networks until we see a decrease in our validation accuracy, and an increase in train accuracy (of at least 99.9\%), we can declare our networks as overfitted. We attempt to correct overfitting by pruning these networks for a better test accuracy than that of the overfitted networks.


\section{Neural Network Compression}
\begin{figure}[b!]
      \centering
      \includegraphics[width=9cm,height=6.5cm]{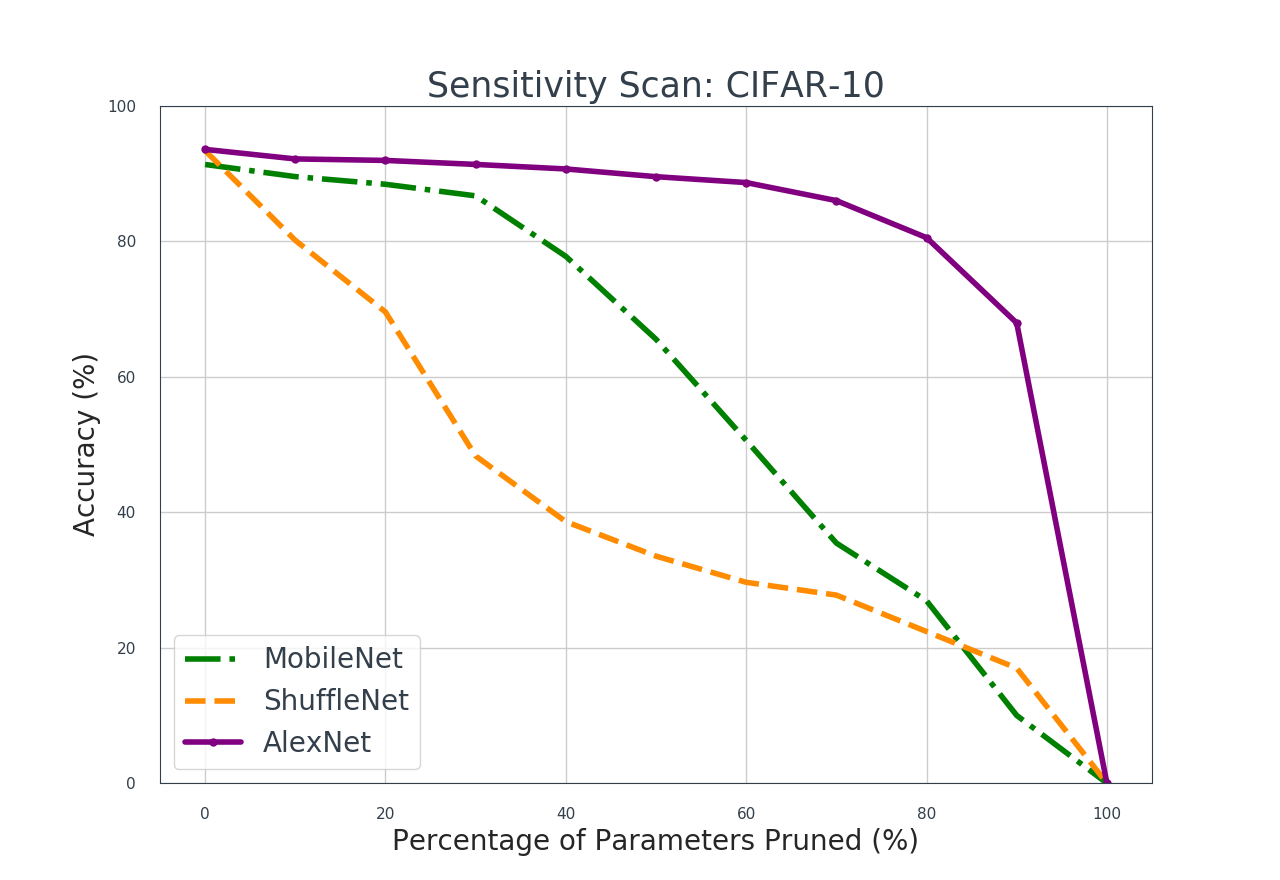}
      \caption{Sensitivity of pruning networks trained on CIFAR-10}
      \label{fig:cifar_sen_scan}
  \end{figure}

\begin{figure}[]
      \centering
      \includegraphics[width=9cm,height=6.5cm]{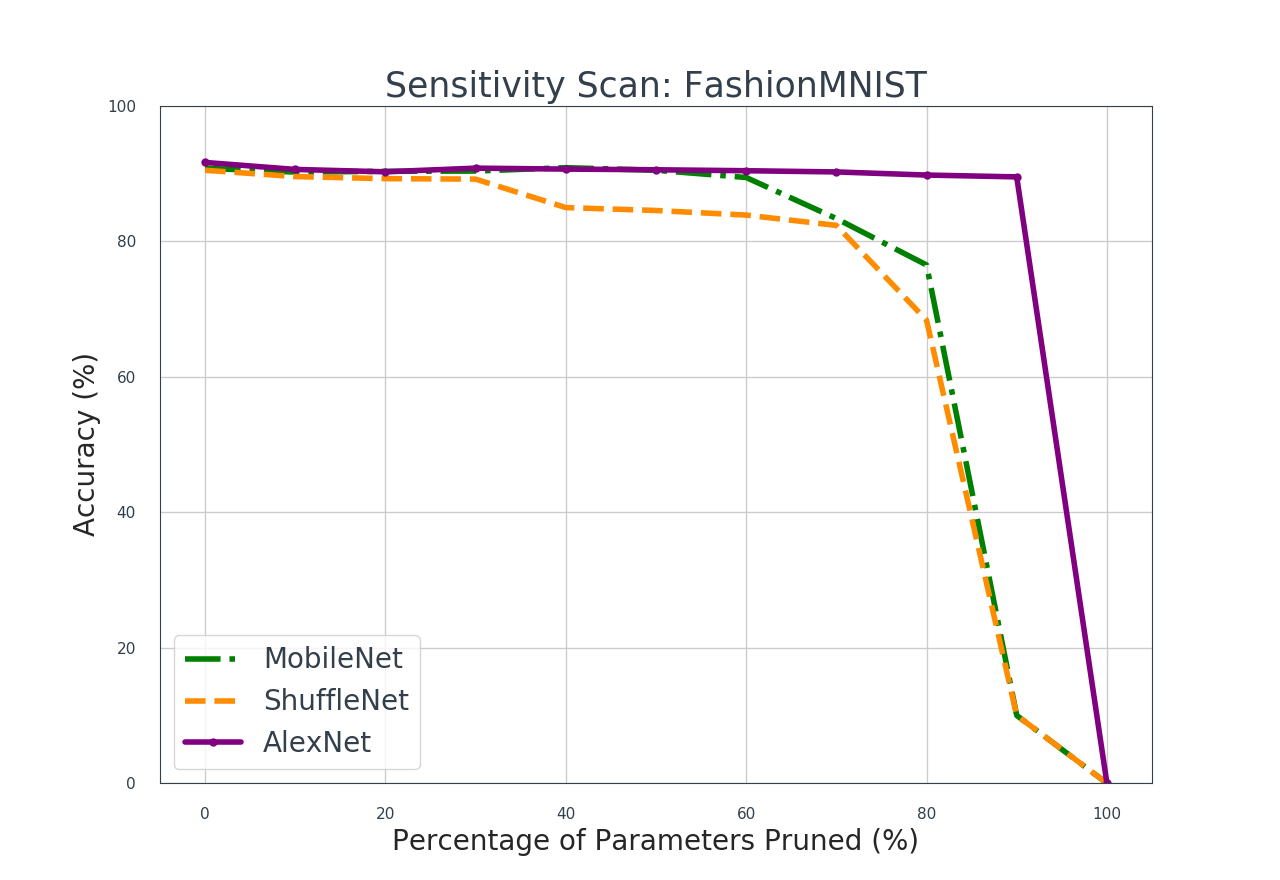}
      \caption{Sensitivity of pruning networks trained on FashionMNIST}
      \label{fig:fmnist_sen_scan}
  \end{figure}

We tested our compression experiments on both the CIFAR-10 and FashionMNIST datasets. The number of parameters to remove from each network were determined through sensitivity scans, as illustrated in Fig. \ref{fig:cifar_sen_scan} (CIFAR-10) and Fig. \ref{fig:fmnist_sen_scan} (FashionMNIST). We note that the sensitivity scans were done to find how many of the smallest weights could be removed without negatively impacting the network's performance, and so the networks were retrained with early-stopping and not fully retrained. 
MobileNet, ShuffleNet and AlexNet performed best with sensitivities of 0.4, 0.3 and 0.5 respectively, on CIFAR-10. With FashionMNIST, MobileNet responded well to a sensitivity of 0.7, while ShuffleNet could be pruned with a sensitivity of 0.35, and AlexNet performed best with a sensitivity of 0.9. These sensitivities were used in the final experiments.

Iteratively pruning each network on both datasets give our results in Table \ref{table:pruned}.
We then quantise the resulting pruned networks to get our final compressed networks shown in Table \ref{table:quantised}.
We retrained the pruned networks with uniformly re-initialising the parameter weights and with fine-tuning (retraining from the pruned parameter weights) the parameters. AlexNet did not respond well to re-initialisation and so the remaining parameters had to be fine-tuned. The compact architectures performed better with re-initialisation compared to fine-tuning, and so we retrained them with uniform re-initialisation.

\begin{table*}[t]
    \centering
    \caption{Network pruning on CIFAR-10 and FashionMNIST}
    \begin{tabular}{@{\extracolsep{4pt}}lcccccc}
        \toprule
        
        \multicolumn{1}{c}{\multirow{2}{*}{\raisebox{-\heavyrulewidth}{\textbf{Network}}}} & \multicolumn{2}{c}{\textbf{Accuracy (\%)}} & \multicolumn{2}{c}{\textbf{Total Parameters}} & \multicolumn{2}{c}{\textbf{Compression Rate}} \\
        \cmidrule{2-3}
        \cmidrule{4-5}
        \cmidrule{6-7}
        {} & CIFAR10 & FashionMNIST & CIFAR10 & FashionMNIST & CIFAR10 & FashionMNIST \\ 
        \midrule
        MobileNet -- Reference  & 91.31 & 90.75 & 2.2M & 2.2M & --- & --- \\
        MobileNet -- Pruned     & 91.53 & 90.43 & 671K & 1.1M & $1.6\times$ & $3.3\times$ \\
        \midrule
        ShuffleNet -- Reference & 93.36 & 90.36 & 1.2M & 1.2M & --- & --- \\
        ShuffleNet -- Pruned    & 93.05 & 90.09 & 879K & 815K & $1.4\times$ & $1.5\times$ \\
        \midrule
        AlexNet -- Reference    & 93.54 & 91.61 & 57M & 57M & --- & --- \\
        AlexNet -- Pruned       & 90.91 & 90.34 & 28M & 5M & $2\times$ & $10\times$ \\
        \bottomrule
    \end{tabular}
    \label{table:pruned}
\end{table*}

\begin{table*}[t]
    \centering
    \caption{Network quantisation of pruned models trained on CIFAR10 and FashionMNIST}
    \begin{tabular}{@{\extracolsep{4pt}}lcccccc}
        \toprule
        
        \multicolumn{1}{c}{\multirow{2}{*}{\raisebox{-\heavyrulewidth}{\textbf{Network}}}} & \multicolumn{2}{c}{\textbf{Accuracy (\%)}} & \multicolumn{2}{c}{\textbf{Size (MB)}} & \multicolumn{2}{c}{\textbf{Inference Time (ms)}} \\
        \cmidrule{2-3}
        \cmidrule{4-5}
        \cmidrule{6-7}
        {} & CIFAR10 & FashionMNIST & CIFAR10 & FashionMNIST & CIFAR10 & FashionMNIST \\ 
        \midrule
        MobileNet -- Reference  & 91.31 & 90.75 & 8.7 & 8.7 & 34.80 & 1.70 \\
        MobileNet -- Quantised  & 90.59 & 90.07 & \textbf{2.9} & \textbf{2.9} & \textbf{4.74} & \textbf{0.30} \\
        \midrule
        ShuffleNet -- Reference & 93.36 & 90.36 & 4.9 & 4.9 & 11.67 & 0.73 \\
        ShuffleNet -- Quantised & 81.29 & 89.78 & \textbf{1.8} & \textbf{1.8} & 23.15 & \textbf{0.61} \\
        \midrule
        AlexNet -- Reference    & 93.54 & 91.61 & 217.6 & 217.6   & 22.13 & 6.70 \\
        AlexNet -- Quantised    & 90.06 & 90.27 & \textbf{54.6} & \textbf{54.6} & \textbf{5.23} & \textbf{4.90} \\
        \bottomrule
    \end{tabular}
    \label{table:quantised}
\end{table*}

We find that MobileNet and ShuffleNet are quite sensitive to pruning as shown in Fig. \ref{fig:cifar_sen_scan}. These compact networks are small and their parameters are less likely to be redundant. AlexNet, on the other hand, is a much larger network and proves to be less sensitive to pruning, indicating many redundant parameters. We removed 90\% of AlexNet's parameters when trained on FashionMNIST and 50\% of its parameters when trained on CIFAR-10, without a significant reduction in accuracy. We find that we get better compression rates on networks trained on FashionMNIST compared to CIFAR-10 largely due to FashionMNIST containing grayscale images of smaller size, and so network complexity can significantly be reduced.
Quantising the pruned versions of MobileNet and AlexNet trained on CIFAR-10 resulted in a considerable reduction of in physical size and inference time, also without a significant loss in accuracy as shown in Table \ref{table:quantised}. Quantisation worked particularly well on MobileNet, leading to a 7.3$\times$ speedup from 34.80ms to 4.74ms on CIFAR-10 and a 5.7$\times$ speedup from 1.70ms to 0.30ms on FashionMNIST. Surprisingly, ShuffleNet trained on CIFAR-10 did not respond well to quantisation: while its size decreases, it suffers a very large accuracy loss with an increase in its inference time. This is possibly due to its more complex architecture and small size, leading quantisation to add more overhead overall.


\section{Neural Network Regularisation}
We trained our networks to overfit FashionMNIST as it required significantly less training time and computational resources compared to CIFAR-10. Once our networks completely overfitted the training data with over 99.9\% training accuracy, we pruned each network until the test accuracy started to decrease. The results are from overfitting and then pruning each network are shown in Table \ref{table:fmnist_overfit}. MobileNet, ShuffleNet and AlexNet were pruned with sensitivities 0.1, 0.15 and 0.6 respectively.

Table \ref{table:fmnist_overfit} demonstrates that pruning is a means to correct overfitting. The parameters pruned away results in a higher test accuracy than that of the overfitted network. This allows the network to generalise better and obtain a higher test accuracy overall. An advantage of using pruning as a regularisation technique is that it can be applied after training, rather than during training as there could be uncertainty as to whether regularisation is needed.

\begin{table}[h!]
    \centering
    \caption{Overfitted and pruned networks trained on FashionMNIST}
    \begin{tabular}{@{\extracolsep{4pt}}lcc}
        \toprule
        
        \multicolumn{1}{c}{\textbf{Network}} & \textbf{Accuracy (\%)} & \textbf{Total Parameters} \\
        \midrule
        MobileNet -- Reference           & 90.75             & 2.2M                    \\ 
        MobileNet -- Overfitted          & 90.55             & 2.2M                    \\ 
        MobileNet -- Pruned              & \textbf{91.26}    & 1.76M                   \\ \midrule 
        ShuffleNet -- Reference          & 90.36             & 1.2M                    \\ 
        ShuffleNet -- Overfitted         & 89.71             & 1.2M                    \\ 
        ShuffleNet -- Pruned             & \textbf{91.01}    & 1M                    \\ \midrule 
        AlexNet -- Reference             & 91.61             & 57M                     \\ 
        AlexNet -- Overfitted            & 91.32             & 57M                     \\ 
        AlexNet -- Pruned                & \textbf{92.11}    & 22M                     \\  
        \bottomrule
    \end{tabular}
    \label{table:fmnist_overfit}
\end{table}


\section{Conclusion}
Across both compact and large networks we demonstrated that pruning is an effective regularisation technique to correct overfitting. We have shown that the compact networks MobileNet and ShuffleNet are still receptive to network pruning as a means of compression and the correction of overfitting, despite having relatively few parameters compared to larger networks such as AlexNet. We found that quantisation significantly decreased the size and computational requirements of AlexNet and MobileNet, but negatively impacted the performance of a more complex ShuffleNet architecture. Compared to a large, compressed network, we find that compact architectures consume less memory and storage, has better accuracy with faster training and inference times, and yet can still benefit from compression techniques with relatively few parameters. Our results suggest that employing compact architectures are more promising to compressing large networks.


\section{Future Work}
This work focused on an iterative pruning method which introduced sparsity into networks. Pruning techniques like filter pruning do not introduce sparsity into the networks allowing them to maintain their original structure, leading to a smaller model size with faster inference times. Filter pruning removes entire filters from convolutional layers but leaves the fully-connected layers untouched. A future direction of research would be to explore the compression rates between these two methods and examine the trade-offs between sparse networks and non-sparse networks.


\bibliography{main}

\end{document}